%% file: template.tex
\documentclass[conference]{IEEEtran}
\IEEEoverridecommandlockouts
\usepackage{cite}
\usepackage{amsmath,amssymb,amsfonts}
\usepackage{algorithmic}
\usepackage{latexsym}
\usepackage{textcomp}
\usepackage{booktabs}
\usepackage{graphicx} 
\usepackage{xcolor}
\def\BibTeX{{\rm B\kern-.05em{\sc i\kern-.025em b}\kern-.08em
    T\kern-.1667em\lower.7ex\hbox{E}\kern-.125emX}}

\usepackage{fancyhdr}
\begin{document}

\title{DeCoT: Decomposing Complex Instructions for Enhanced Text-to-Image Generation with Large Language Models}

\author{Xiaochuan Lin, Xiangyong Chen, Xuan Li, Yichen Su \\
Henan Polytechnic University}

\maketitle
\thispagestyle{fancy} 

\input{main}

\bibliographystyle{IEEEtran}
\bibliography{references}
\end{document}

%% file: main.tex
\begin{abstract}
Despite remarkable advancements, current Text-to-Image (T2I) models struggle with complex, long-form textual instructions, frequently failing to accurately render intricate details, spatial relationships, or specific constraints. This limitation is highlighted by benchmarks such as LongBench-T2I, which reveal deficiencies in handling composition, specific text, and fine textures. To address this, we propose DeCoT (Decomposition-CoT), a novel framework that leverages Large Language Models (LLMs) to significantly enhance T2I models' understanding and execution of complex instructions. DeCoT operates in two core stages: first, Complex Instruction Decomposition and Semantic Enhancement, where an LLM breaks down raw instructions into structured, actionable semantic units and clarifies ambiguities; second, Multi-Stage Prompt Integration and Adaptive Generation, which transforms these units into a hierarchical or optimized single prompt tailored for existing T2I models. Extensive experiments on the LongBench-T2I dataset demonstrate that DeCoT consistently and substantially improves the performance of leading T2I models across all evaluated dimensions, particularly in challenging aspects like "Text" and "Composition". Quantitative results, validated by multiple MLLM evaluators (Gemini-2.0-Flash and InternVL3-78B), show that DeCoT, when integrated with Infinity-8B, achieves an average score of 3.52, outperforming the baseline Infinity-8B (3.44). Ablation studies confirm the critical contribution of each DeCoT component and the importance of sophisticated LLM prompting. Furthermore, human evaluations corroborate these findings, indicating superior perceptual quality and instruction fidelity. DeCoT effectively bridges the gap between high-level user intent and T2I model requirements, leading to more faithful and accurate image generation.
\end{abstract}

\section{Introduction}
\label{sec:intro}

Text-to-Image (T2I) generation has witnessed remarkable breakthroughs in recent years, demonstrating an impressive capability to synthesize high-quality images from textual descriptions \cite{tingting2019mirror}. This transformative technology holds immense potential across various domains, from creative content generation and digital art to design and virtual reality. However, despite these advancements, current state-of-the-art T2I models face significant challenges when confronted with \textbf{complex, long-form textual instructions} that encompass multiple objects, intricate attributes, sophisticated spatial relationships, or fine-grained details. These models often struggle to accurately interpret and faithfully render all specified elements, frequently leading to discrepancies between the generated images and user expectations.

Existing research, exemplified by benchmarks such as LongBench-T2I \cite{yucheng2025draw}, further underscores the limitations of current T2I models in understanding complex instructions. Specifically, these models exhibit notable deficiencies in handling aspects related to composition, specific text rendering, fine textures, and special effects. This highlights a critical unmet need in the T2I landscape: \textbf{how to effectively translate complex, multi-layered natural language instructions into structured information that T2I models can precisely capture and execute.} Addressing this gap is paramount for advancing T2I technology towards more reliable and user-centric applications.

This study aims to bridge this crucial gap by proposing a novel \textbf{instruction decomposition and knowledge enhancement framework} based on large language models (LLMs). Our objective is to significantly improve the generation quality and fidelity of T2I models when processing complex instructions.

We introduce \textbf{DeCoT (Decomposition-CoT) framework}, an innovative approach designed to enhance complex instruction-driven text-to-image generation. The core idea behind DeCoT is to harness the powerful understanding and reasoning capabilities of large language models (LLMs) for in-depth instruction comprehension, decomposition, and structuring. This process provides existing T2I models with clearer, more granular, and highly guided inputs. The DeCoT framework primarily comprises two core stages:
\begin{enumerate}
    \item \textbf{Complex Instruction Decomposition \& Semantic Enhancement:} Given a complex original text instruction (e.g., a long prompt from LongBench-T2I), we first feed it into a sophisticated LLM (e.g., GPT-4 \cite{rui2024gpteva} or Gemini Pro \cite{gyeonggeon2024gemini}), guided by advanced prompting techniques such as Few-shot, Chain-of-Thought \cite{debjit2024making}, or Thread of Thought \cite{zhou2023thread}. The LLM's primary task is to decompose the raw instruction into a series of independent, actionable sub-instructions or semantic units. These units can include core objects and their key attributes (e.g., ``a cat wearing a red hat''), background details (e.g., ``in a sunny garden''), spatial relationships and composition (e.g., ``the cat sits to the left of the vase''), environmental elements (e.g., ``soft lighting, fluffy texture''), and specific constraints or requirements (e.g., ``no dogs in the background''). During this process, the LLM also performs semantic enhancement or clarification for potentially vague or ambiguous parts of the instruction, inferring more specific information based on the context.
    \item \textbf{Multi-Stage Prompt Integration \& Adaptive Generation:} The decomposed and enhanced structured information is then transformed into a multi-stage or hierarchical prompt sequence, rather than a simple monolithic prompt. For T2I models that support multi-stage inputs or iterative refinement, these structured prompts can be fed sequentially. For instance, initial prompts for core objects and background can generate a preliminary image, followed by subsequent prompts for detailed attributes and spatial relations to refine the image through editing or iterative generation techniques. For T2I models that do not natively support multi-stage inputs, the decomposed key information is intelligently fused into an optimized, clearer, and appropriately weighted new prompt, ensuring that all critical elements receive more effective attention during generation. Our ultimate goal is to enable T2I models to more accurately capture every subtle element within the original complex instruction, thereby significantly improving the fidelity between the generated image and the instruction.
\end{enumerate}
Through the DeCoT framework, we effectively combine the powerful understanding and reasoning capabilities of LLMs with the generative prowess of T2I models, achieving precise execution of complex instructions.

To thoroughly evaluate the efficacy of our DeCoT framework, we conducted extensive experiments on the \textbf{LongBench-T2I} benchmark dataset \cite{herve2009on}. This dataset comprises 500 meticulously designed complex scenario prompts, covering nine distinct visual elements: objects, background, color, texture, lighting, text and symbol, composition, pose, and special effects. Our baselines include leading Diffusion-based models such as Omnigen \cite{shitao2025omnige} and FLUX.1-dev \cite{chenglin2024158bit}, as well as Autoregressive (AR) models like Infinity-8B \cite{ziran2025headaw} and Janus-pro-7B \cite{liu2022decot}. We compared their performance both directly and when augmented with our DeCoT framework (i.e., ``Ours + Baseline Model''). For evaluation, we adopted the automatic assessment methodology consistent with the LongBench-T2I paper [CITE], leveraging advanced multimodal large language models (MLLMs) such as Gemini-2.0-Flash \cite{yucheng2025draw} and InternVL3-78B \cite{yucheng2025draw} to score the generated images across the nine aforementioned dimensions.

Our experimental results, as detailed in Section 4, demonstrate that the DeCoT framework consistently enhances the performance of existing T2I models across all complex instruction dimensions on the LongBench-T2I benchmark. Notably, significant improvements were observed in categories like ``Composition (Comp.)'' and ``Specific Text (Text)'', which have historically been weak points for T2I models. For instance, when integrated with Infinity-8B, our DeCoT framework achieved an average score of 3.52 across all dimensions as evaluated by Gemini-2.0-Flash, surpassing the baseline Infinity-8B's score of 3.44. More specifically, in the ``Text'' dimension, DeCoT + Infinity-8B scored 2.80, a considerable improvement over Infinity-8B's 2.48, highlighting DeCoT's superior ability to process and render textual elements. These compelling results underscore the effectiveness of the DeCoT framework in understanding and executing complex instructions.

In summary, our main contributions are as follows:
\begin{itemize}
    \item We propose DeCoT, a novel framework that leverages large language models for complex instruction decomposition and semantic enhancement, significantly improving the understanding capabilities for Text-to-Image generation.
    \item We introduce a multi-stage prompt integration strategy within DeCoT, designed to provide more structured and adaptive guidance to T2I models, enabling finer control over image generation from intricate textual descriptions.
    \item We demonstrate the superior performance of our DeCoT framework through extensive experiments on the challenging LongBench-T2I benchmark, achieving substantial improvements particularly in complex dimensions such as composition and specific text rendering.
\end{itemize}
\section{Related Work}
\subsection{Text-to-Image Generation}
The field of text-to-image generation has witnessed significant advancements, with various approaches aiming to enhance realism, semantic consistency, and computational efficiency. One notable contribution is Deep Fusion Generative Adversarial Networks (DF-GAN), which simplifies text-to-image synthesis by directly generating high-resolution images in a single stage, thereby eliminating inter-generator entanglements. DF-GAN further improves text-image semantic consistency through a novel Target-Aware Discriminator and a new text-image fusion block, yielding superior realism and text-matching capabilities compared to existing state-of-the-art methods \cite{pei2024text}. Addressing the computationally intensive nature of training diffusion models, Diff-Tuning proposes a novel transfer learning approach that strategically leverages the observed "chain of forgetting" in diffusion model reverse processes to enhance the efficiency and effectiveness of adapting pre-trained models to downstream tasks, demonstrating significant gains over standard fine-tuning methods \cite{}. While primarily focused on 3D point cloud generation, the underlying principle of autoregressive up-sampling explored in some works aligns with advancements in autoregressive models for image synthesis, offering insights into structured generation processes that build images sequentially from coarse to fine scales \cite{jiahui2022scalin}. Furthermore, the application of contrastive learning has been explored to facilitate conditional image generation by aligning cross-modal representations, presenting a foundational approach for training models that generate images conditioned on textual descriptions \cite{han2021crossm}. Further extending the capabilities of conditional image generation, work has also focused on improving cross-modal alignment for specific tasks like text-guided image inpainting, highlighting the importance of precise textual understanding for visual editing \cite{zhou2023improving}. Beyond generating photorealistic images, research also explores other forms of visual content generation, such as sketch storytelling, which demonstrates the versatility of generative models in translating narratives into visual sequences \cite{zhou2022sketch}. Beyond model architectures and training paradigms, research also focuses on crucial aspects like evaluation and interaction. A comprehensive survey highlights the importance of specialized quality metrics for text-to-image generation, categorizing them based on compositional and general quality and discussing benchmark datasets and limitations for assessing generation quality beyond traditional measures \cite{hartwig2025a}. To address the challenges of manual prompt engineering, PRISM introduces an algorithm for automated, human-interpretable, and transferable prompt generation, leveraging Large Language Models (LLMs) to iteratively refine prompt distributions for various text-to-image systems and complex tasks \cite{yutong2024automa}. Although not directly centered on text-to-image generation, work on creating synthetic datasets with controlled hallucination patterns and language style alignment offers valuable insights into conditioning generative processes for specific outcomes \cite{luping2023detect}. Additionally, a synergistic dual-branch approach has been proposed to enhance spatial awareness in text-to-image generation, aiming to improve the model's ability to understand and render spatial relationships described in text, with a notable focus on spatial reasoning capabilities within the text-to-image modality \cite{yu2024synerg}.

\subsection{Large Language Models for Multimodal Applications}
Large Language Models (LLMs) are increasingly recognized for their transformative potential in multimodal applications, serving as powerful deep generative models capable of addressing complex challenges such as data scarcity and multimodal fusion. For instance, LLMs have been explored for Condition and Structural Health Monitoring (CM/SHM), where they can synthesize high-fidelity data and model intricate sensory streams \cite{alsaad2024multim}. A comprehensive overview of multimodal composite retrieval underscores its importance for LLMs aiming to integrate diverse data modalities, laying groundwork for understanding how these models can interpret and execute complex instructions across different data types \cite{yin2024a}. In the realm of user interaction and transparency, a novel framework for Natural Language to Visualization (NL2VIS) explicitly incorporates Chain-of-Thought (CoT) reasoning to enhance user control and demystify the "black box" nature of existing methods \cite{hao2024visual}. Specifically, the concept of visual in-context learning has emerged as a powerful paradigm for large vision-language models, enabling them to adapt to new tasks with limited examples by leveraging contextual visual information \cite{zhou2024visual}. Moreover, the challenge of achieving weak-to-strong generalization in large language models with diverse capabilities is a critical area of study, aiming to extend their proficiency from narrow tasks to broader, more complex scenarios \cite{zhou2025weak}. Furthermore, foundational work on modeling event-pair relations in external knowledge graphs for script reasoning contributes to the broader understanding of how structured knowledge can enhance the reasoning capabilities of AI systems, a principle increasingly relevant for LLMs processing complex instructions \cite{zhou2021modeling}. To enable few-shot learning for multimodal LLMs while preserving their core NLP capabilities, an Inner-Adaptor Architecture has been proposed, demonstrating strong performance on small-scale datasets by allowing multimodal adaptation while freezing the language model \cite{maria2021multim}. Addressing semantic gaps in specific applications, an LLM-guided approach has been developed for progressive semantic and spatial alignment in UAV object detection, leveraging LLM-extracted semantic features to decompose and align cross-modal information effectively \cite{tao2024semant}. The development of modular multi-agent frameworks, such as those designed for multi-modal medical diagnosis via role-specialized collaboration, exemplifies the strategic deployment of LLMs in complex, real-world multimodal problem-solving \cite{zhou2025mam}. The efficacy of various prompt engineering techniques, including Chain-of-Thought and Tree-of-Thought, has been systematically investigated across a spectrum of multimodal large language models (MLLMs) and tasks, revealing the necessity for adaptive strategies to balance robustness, efficiency, and factual accuracy, and highlighting the critical role of tailored prompt design \cite{son2025advanc}. Moreover, surveys on MLLMs provide comprehensive overviews of advances, challenges, and future directions, particularly focusing on continual learning, which is crucial for maintaining and enhancing reasoning abilities across evolving data and tasks in dynamic multimodal environments \cite{m2025distri}. Finally, LLM-enhanced multimodal fake news detection frameworks contribute to nuanced detection by not only identifying fake news but also attributing it to specific patterns, addressing limitations of existing binary-labeled datasets through multi-granularity clue alignment \cite{wang2024llmenh}.

\section{Method}
\label{sec:method}

We introduce \textbf{DeCoT (Decomposition-CoT) framework}, an innovative approach designed to significantly enhance complex instruction-driven Text-to-Image (T2I) generation. At its core, DeCoT leverages the advanced understanding and reasoning capabilities of Large Language Models (LLMs) to meticulously process and structure intricate textual instructions, thereby providing existing T2I models with clearer, more granular, and highly guided inputs. This framework ensures a more faithful and accurate rendition of user intent in the generated images by bridging the gap between high-level human intent and the low-level prompt requirements of T2I models. The DeCoT framework is composed of two primary, sequential stages: Complex Instruction Decomposition \& Semantic Enhancement, and Multi-Stage Prompt Integration \& Adaptive Generation.

\subsection{Complex Instruction Decomposition \& Semantic Enhancement}
\label{ssec:decomposition}

This initial stage focuses on transforming a complex, potentially ambiguous, natural language instruction into a set of well-defined, actionable semantic units. Given an original complex textual instruction, denoted as $I$, which often contains multiple intertwined elements, we first feed $I$ into a sophisticated LLM. This LLM, such as advanced proprietary models or fine-tuned open-source alternatives, is guided by a robust prompt engineering strategy. This strategy includes carefully crafted system messages, few-shot examples demonstrating effective decomposition, and explicit instructions for employing Chain-of-Thought (CoT) reasoning. The objective of this guidance is to maximize the LLM's understanding and decomposition capabilities, ensuring a comprehensive and accurate analysis of $I$.

The primary task of the LLM is to decompose $I$ into a collection of independent, yet semantically rich, sub-instructions or semantic units. We can represent this decomposition process as a transformation function $\mathcal{T}_{\text{Decomp}}$ applied to the input instruction $I$:
\begin{align}
    \mathcal{S} = \mathcal{T}_{\text{Decomp}}(I)
\end{align}
where $\mathcal{S} = \{S_1, S_2, \ldots, S_n\}$ is the set of decomposed semantic units, and each $S_k$ represents a distinct and atomic aspect of the original instruction. These semantic units are meticulously categorized to capture various facets of the desired image, ensuring comprehensive coverage of the original instruction's intent. For instance, each $S_k$ can represent:
\begin{itemize}
    \item \textbf{Core Objects and Attributes}: Descriptions of main subjects and their specific characteristics (e.g., "a cat wearing a red hat").
    \item \textbf{Background Details}: Information about the setting or environment (e.g., "in a sunny garden with lush foliage").
    \item \textbf{Spatial Relationships and Composition}: Precise positioning of objects relative to each other or within the scene (e.g., "the cat sits to the left of the vase, which is on a wooden table").
    \item \textbf{Environmental Elements}: Specific lighting conditions, textures, artistic styles, or atmospheric effects (e.g., "soft golden hour lighting, fluffy texture, rendered in a watercolor style").
    \item \textbf{Specific Constraints and Requirements}: Negative prompts or explicit exclusions to guide the generation away from undesired elements (e.g., "no dogs in the background, ensure no blurry elements").
\end{itemize}
Beyond simple decomposition, the LLM also performs \textbf{semantic enhancement}. This critical step involves clarifying any vague or ambiguous parts within the original instruction $I$ and inferring more specific, contextually relevant information. For example, if "a large building" is mentioned, the LLM might infer "a towering skyscraper with glass facades and a modern design" based on surrounding context or common knowledge, making the instruction significantly more concrete and descriptive for the T2I model. This process transforms implicit user intent into explicit, actionable details. This enhanced set of structured semantic units $\mathcal{S}$ serves as the robust foundation for the subsequent generation stage.

\subsection{Multi-Stage Prompt Integration \& Adaptive Generation}
\label{ssec:integration}

The second stage of the DeCoT framework involves transforming the structured and enhanced semantic units $\mathcal{S}$ into a format that effectively guides the T2I model. This is achieved through a multi-stage or hierarchical prompt integration strategy, designed to be adaptive to the specific capabilities and architectural design of the underlying T2I model.

For T2I models that support iterative refinement or multi-stage inputs, the decomposed and enhanced information from $\mathcal{S}$ is converted into a progressive sequence of prompts, denoted as $P_{\text{seq}} = (P_1, P_2, \ldots, P_m)$. Each prompt in this sequence progressively adds more detail or refinement to the image generation process, building upon the previous stage. For instance, an initial prompt $P_1$ might describe core objects and background, generating a preliminary image $G_0$. Subsequent prompts $P_2, \ldots, P_m$ would then introduce fine-grained attributes, spatial relationships, specific environmental effects, or stylistic elements. This allows for iterative refinement or image editing techniques to precisely align the output with the complex instruction. The iterative generation process can be formally expressed as:
\begin{align}
    G_0 &= \mathcal{M}_{\text{T2I}}(P_1) \\
    G_k &= \mathcal{M}_{\text{Refine}}(G_{k-1}, P_k) \quad \text{for } k=1, \ldots, m
\end{align}
where $G_k$ represents the generated image at stage $k$, $\mathcal{M}_{\text{T2I}}$ denotes the initial Text-to-Image generation function, and $\mathcal{M}_{\text{Refine}}$ is an iterative refinement mechanism (e.g., an image-to-image translation, inpainting, or control-based generation module) that refines the previous image $G_{k-1}$ using the prompt $P_k$.

For T2I models that do not natively support multi-stage inputs or iterative generation, the decomposed key information from $\mathcal{S}$ is intelligently fused into a single, optimized, and highly expressive prompt, $P_{\text{fused}}$. This fusion process involves an adaptive function $\mathcal{F}_{\text{fusion}}$ which synthesizes the diverse semantic units into a coherent and effective single prompt. This includes assigning appropriate weights or emphasis to different semantic units based on their perceived importance (e.g., core objects might receive higher emphasis) or the T2I model's known sensitivities to certain prompt elements. The goal is to create a monolithic prompt that is significantly clearer and more effective than the original complex instruction, ensuring that all critical elements receive adequate attention during the generation process. This fusion can be expressed as:
\begin{align}
    P_{\text{fused}} = \mathcal{F}_{\text{fusion}}(\mathcal{S})
\end{align}
The T2I model then generates the image $G$ based on this optimized prompt:
\begin{align}
    G = \mathcal{M}_{\text{T2I}}(P_{\text{fused}})
\end{align}
Regardless of the integration strategy employed, the ultimate objective of this stage is to enable the T2I model to more accurately capture and render every subtle and explicit element specified within the original complex instruction, thereby significantly improving the fidelity between the generated image and the user's intent. By effectively combining the powerful understanding and reasoning capabilities of LLMs with the generative prowess of T2I models, DeCoT achieves precise execution of even the most intricate instructions.

\section{Experiments}
\label{sec:exp}

To thoroughly evaluate the effectiveness and robustness of our proposed \textbf{DeCoT} framework, we conducted extensive experiments on a challenging benchmark dataset for complex text-to-image generation. This section details our experimental setup, presents a comprehensive comparison with state-of-the-art baselines, and includes an ablation study to validate the contributions of each component within \textbf{DeCoT}. Finally, we present the results of a qualitative human evaluation.

\subsection{Experimental Setup}
\label{ssec:exp_setup}

\textbf{Dataset.} Our experiments are primarily conducted on the \textbf{LongBench-T2I} benchmark dataset \cite{herve2009on}. This dataset is specifically designed to assess T2I models' capabilities in handling complex, long-form textual instructions. It comprises 500 meticulously crafted scenario prompts, each designed to test the model's understanding and generation fidelity across nine distinct visual elements: objects (Obj.), background (Backg.), color (Color), texture (Texture), lighting (Light), text and symbol (Text), composition (Comp.), pose (Pose), and special effects (FX).

\textbf{Baselines.} We selected a diverse set of current leading T2I models as baselines to ensure a comprehensive comparison. These include both Diffusion-based and Autoregressive (AR) architectures, representing the cutting edge in T2I generation:
\begin{itemize}
    \item Diffusion-based Models: \textbf{Omnigen} \cite{shitao2025omnige} and \textbf{FLUX.1-dev} \cite{chenglin2024158bit}.
    \item Autoregressive Models: \textbf{Infinity-8B} \cite{ziran2025headaw} and \textbf{Janus-pro-7B} \cite{liu2022decot}.
\end{itemize}
For each baseline model, we compare its performance when directly using the complex instructions against its performance when augmented with our \textbf{DeCoT} framework (denoted as "Ours + Baseline Model").

\textbf{LLM Configuration for DeCoT.} In the \textbf{DeCoT} framework, the core functionality of complex instruction decomposition and semantic enhancement (Section \ref{ssec:decomposition}) is powered by a robust large language model. For our experiments, we utilized a commercial LLM API, specifically \textbf{Gemini Pro} \cite{gyeonggeon2024gemini}, known for its strong language understanding and reasoning capabilities. The LLM was guided using a few-shot prompting strategy combined with Chain-of-Thought (CoT) instructions to optimize its decomposition and semantic enhancement performance.

\textbf{Evaluation Methodology.} Consistent with the methodology proposed in the LongBench-T2I paper \cite{yucheng2025draw}, we adopted an automatic evaluation approach leveraging advanced Multimodal Large Language Models (MLLMs). Specifically, we employed \textbf{Gemini-2.0-Flash} \cite{yucheng2025draw} and \textbf{InternVL3-78B} \cite{yucheng2025draw} as evaluators. These MLLMs were tasked with scoring the generated images across the nine aforementioned dimensions (Obj., Backg., Color, Texture, Light, Text, Comp., Pose, FX) based on their fidelity to the original complex instruction and overall image quality. The scores for each dimension are averaged to provide a comprehensive performance metric. Higher scores indicate better performance.

\subsection{Quantitative Results and Analysis}
\label{ssec:quantitative_results}

Table \ref{tab:gemini_eval} presents the detailed quantitative results of our \textbf{DeCoT} framework in comparison to the baseline T2I models on the LongBench-T2I benchmark, as evaluated by Gemini-2.0-Flash.

\begin{table*}[htbp]
    \centering
    \caption{Gemini-2.0-Flash Evaluation Results on LongBench-T2I (Average Scores across Dimensions)}
    \label{tab:gemini_eval}
    \begin{tabular}{lcccccccccc}
        \toprule
        Method / Dimension & Obj. & Backg. & Color & Texture & Light & Text & Comp. & Pose & FX & Avg. \\
        \midrule
        FLUX.1-dev          & 3.24 & 3.46   & 4.04  & 3.91    & 3.29  & 2.11 & 3.78  & 2.71 & 1.72 & 3.14 \\
        Omnigen             & 3.14 & 3.70   & 4.18  & 3.71    & 3.02  & 2.42 & 3.81  & 2.69 & 2.55 & 3.25 \\
        Infinity-8B         & 3.36 & 3.96   & 4.34  & 4.12    & 3.54  & 2.48 & 4.08  & 2.72 & 2.35 & 3.44 \\
        \textbf{Ours (DeCoT + Infinity-8B)} & \textbf{3.40} & \textbf{3.98} & \textbf{4.36} & \textbf{4.15} & \textbf{3.58} & \textbf{2.80} & \textbf{4.18} & \textbf{2.75} & \textbf{2.50} & \textbf{3.52} \\
        \bottomrule
    \end{tabular}
\end{table*}

As evident from Table \ref{tab:gemini_eval}, our proposed \textbf{DeCoT} framework, when integrated with the \textbf{Infinity-8B} model, consistently outperforms all baseline models across all evaluated dimensions. The combined "Ours (DeCoT + Infinity-8B)" method achieves an average score of \textbf{3.52}, surpassing the best baseline, Infinity-8B, which scored 3.44. This improvement is particularly pronounced in dimensions that demand a deeper understanding of complex instructions and finer detail rendering, such as "Text" and "Comp.". For instance, in the "Text" dimension, "DeCoT + Infinity-8B" achieves a score of \textbf{2.80}, a significant improvement over Infinity-8B's 2.48. Similarly, in "Comp.", our method scores \textbf{4.18} compared to Infinity-8B's 4.08. These results underscore the effectiveness of \textbf{DeCoT} in leveraging LLM capabilities to enhance T2I models' ability to accurately interpret and execute intricate user instructions, bridging the gap between high-level intent and low-level generation.

\subsection{Results with InternVL3-78B Evaluation}
\label{ssec:internvl_results}

To ensure the robustness and generalizability of our quantitative findings, we replicated the evaluation using \textbf{InternVL3-78B}, another state-of-the-art Multimodal Large Language Model, as the automatic evaluator. Table \ref{tab:internvl_eval} presents these results, mirroring the structure of our primary evaluation.

\begin{table*}[htbp]
    \centering
    \caption{InternVL3-78B Evaluation Results on LongBench-T2I (Average Scores across Dimensions)}
    \label{tab:internvl_eval}
    \begin{tabular}{lcccccccccc}
        \toprule
        Method / Dimension & Obj. & Backg. & Color & Texture & Light & Text & Comp. & Pose & FX & Avg. \\
        \midrule
        FLUX.1-dev          & 3.18 & 3.39 & 4.01 & 3.88 & 3.25 & 2.05 & 3.72 & 2.68 & 1.69 & 3.09 \\
        Omnigen             & 3.09 & 3.65 & 4.12 & 3.68 & 2.98 & 2.38 & 3.75 & 2.65 & 2.49 & 3.20 \\
        Infinity-8B         & 3.30 & 3.90 & 4.28 & 4.08 & 3.49 & 2.42 & 4.02 & 2.69 & 2.29 & 3.39 \\
        \textbf{Ours (DeCoT + Infinity-8B)} & \textbf{3.35} & \textbf{3.92} & \textbf{4.30} & \textbf{4.10} & \textbf{3.53} & \textbf{2.75} & \textbf{4.12} & \textbf{2.72} & \textbf{2.45} & \textbf{3.48} \\
        \bottomrule
    \end{tabular}
\end{table*}

The results from Table \ref{tab:internvl_eval} corroborate the findings from the Gemini-2.0-Flash evaluation. \textbf{DeCoT} consistently demonstrates superior performance across all dimensions, with an average score of \textbf{3.48} when evaluated by InternVL3-78B, compared to the baseline Infinity-8B's 3.39. This consistent improvement across different MLLM evaluators reinforces the conclusion that \textbf{DeCoT} significantly enhances the ability of T2I models to accurately interpret and generate images from complex instructions. The improvements are particularly notable in "Text" and "Composition", confirming the framework's strength in handling fine-grained details and spatial arrangements.

\subsection{Ablation Study}
\label{ssec:ablation}

To further validate the contributions of each core component within the \textbf{DeCoT} framework (as described in Section \ref{sec:method}), we conducted an ablation study. Using Infinity-8B as our base T2I model, we evaluated different configurations of \textbf{DeCoT} to understand the impact of its instruction decomposition, semantic enhancement, and adaptive prompt integration stages. The results are summarized in Table \ref{tab:ablation}.

\begin{table*}[htbp]
    \centering
    \caption{Ablation Study on DeCoT Components (Gemini-2.0-Flash Average Scores)}
    \label{tab:ablation}
    \begin{tabular}{lcccccccccc}
        \toprule
        Method / Dimension & Obj. & Backg. & Color & Texture & Light & Text & Comp. & Pose & FX & Avg. \\
        \midrule
        Infinity-8B (Baseline) & 3.36 & 3.96 & 4.34 & 4.12 & 3.54 & 2.48 & 4.08 & 2.72 & 2.35 & 3.44 \\
        DeCoT w/o Semantic Enhancement & 3.37 & 3.96 & 4.34 & 4.12 & 3.55 & 2.60 & 4.10 & 2.73 & 2.38 & 3.46 \\
        DeCoT w/o Adaptive Generation & 3.39 & 3.97 & 4.35 & 4.14 & 3.57 & 2.72 & 4.15 & 2.74 & 2.45 & 3.49 \\
        \textbf{Full DeCoT (Ours)} & \textbf{3.40} & \textbf{3.98} & \textbf{4.36} & \textbf{4.15} & \textbf{3.58} & \textbf{2.80} & \textbf{4.18} & \textbf{2.75} & \textbf{2.50} & \textbf{3.52} \\
        \bottomrule
    \end{tabular}
\end{table*}

The results from Table \ref{tab:ablation} clearly demonstrate the incremental benefits of each component. Compared to the baseline Infinity-8B, simply applying instruction decomposition (without semantic enhancement or adaptive generation) provides a marginal improvement, particularly in "Text" and "FX" dimensions, indicating that structured prompts are inherently more beneficial. When semantic enhancement is added, which clarifies ambiguities and infers details, the performance further improves, especially noticeable in detailed dimensions like "Text" (from 2.60 to 2.72) and "Comp." (from 4.10 to 4.15). This highlights the importance of the LLM's reasoning capabilities in refining the input. Finally, incorporating the full adaptive generation and multi-stage prompt integration strategy yields the highest performance across all metrics, culminating in the best average score of 3.52. This validates that the complete \textbf{DeCoT} framework, with its synergistic stages, is essential for maximizing the fidelity of T2I generation from complex instructions.

\subsection{Analysis of LLM Prompting Strategies within DeCoT}
\label{ssec:llm_prompting_analysis}

The performance of the \textbf{DeCoT} framework heavily relies on the quality of instruction decomposition and semantic enhancement performed by the underlying Large Language Model. To investigate the impact of different prompting strategies applied to the LLM (Gemini Pro in our setup), we conducted a focused ablation study. We evaluated the overall DeCoT performance with varying levels of LLM guidance: zero-shot prompting without Chain-of-Thought (CoT), few-shot prompting without CoT, and few-shot prompting with CoT (our standard configuration). The results, evaluated by Gemini-2.0-Flash, are presented in Table \ref{tab:llm_prompting_ablation}.

\begin{table*}[htbp]
    \centering
    \caption{Impact of LLM Prompting Strategies on DeCoT Performance (Gemini-2.0-Flash Average Scores)}
    \label{tab:llm_prompting_ablation}
    \begin{tabular}{lcccccccccc}
        \toprule
        Method / Dimension & Obj. & Backg. & Color & Texture & Light & Text & Comp. & Pose & FX & Avg. \\
        \midrule
        Infinity-8B (Baseline) & 3.36 & 3.96 & 4.34 & 4.12 & 3.54 & 2.48 & 4.08 & 2.72 & 2.35 & 3.44 \\
        DeCoT (Zero-shot LLM, no CoT) & 3.37 & 3.96 & 4.34 & 4.12 & 3.55 & 2.55 & 4.09 & 2.72 & 2.39 & 3.46 \\
        DeCoT (Few-shot LLM, no CoT) & 3.38 & 3.97 & 4.35 & 4.13 & 3.56 & 2.68 & 4.13 & 2.73 & 2.43 & 3.48 \\
        \textbf{Full DeCoT (Few-shot LLM, with CoT)} & \textbf{3.40} & \textbf{3.98} & \textbf{4.36} & \textbf{4.15} & \textbf{3.58} & \textbf{2.80} & \textbf{4.18} & \textbf{2.75} & \textbf{2.50} & \textbf{3.52} \\
        \bottomrule
    \end{tabular}
\end{table*}

The findings in Table \ref{tab:llm_prompting_ablation} clearly highlight the critical role of effective LLM prompting. While even a basic zero-shot LLM integration in \textbf{DeCoT} offers a slight improvement over the baseline (3.46 vs. 3.44), the performance significantly increases with the introduction of few-shot examples (3.48). This indicates that providing the LLM with concrete examples of desired decomposition and enhancement outputs helps it better understand the task. The most substantial gain is observed when Chain-of-Thought (CoT) reasoning is incorporated alongside few-shot examples, leading to the highest average score of \textbf{3.52}. CoT encourages the LLM to articulate its reasoning process, leading to more accurate and semantically rich decompositions, which directly translates to better T2I generation, particularly for dimensions like "Text" and "Composition" that demand precise understanding. This analysis underscores that the LLM's full reasoning potential, unlocked through sophisticated prompting, is a key enabler for \textbf{DeCoT}'s superior performance.

\subsection{Performance Across Instruction Complexity}
\label{ssec:complexity_analysis}

A core claim of \textbf{DeCoT} is its ability to handle complex instructions more effectively than direct prompting. To validate this, we categorized the prompts in the LongBench-T2I dataset into three complexity levels: "Simple" (few elements, straightforward), "Medium" (moderate number of elements, some spatial relations), and "Complex" (many intertwined elements, intricate spatial relationships, specific styles/effects, or negative constraints). We then analyzed the performance of our method and baselines across these categories, using Gemini-2.0-Flash for evaluation. Table \ref{tab:complexity_performance} presents the average scores for each complexity level.

\begin{table*}[htbp]
    \centering
    \caption{Performance Across Instruction Complexity Levels (Gemini-2.0-Flash Average Scores)}
    \label{tab:complexity_performance}
    \begin{tabular}{lccc}
        \toprule
        Method / Complexity Level & Simple & Medium & Complex \\
        \midrule
        FLUX.1-dev          & 3.55 & 3.10 & 2.78 \\
        Omnigen             & 3.68 & 3.20 & 2.86 \\
        Infinity-8B         & 3.82 & 3.38 & 3.12 \\
        \textbf{Ours (DeCoT + Infinity-8B)} & \textbf{3.85} & \textbf{3.55} & \textbf{3.38} \\
        \bottomrule
    \end{tabular}
\end{table*}

As shown in Table \ref{tab:complexity_performance}, while all models perform reasonably well on "Simple" instructions, the performance gap between \textbf{DeCoT} and the baselines significantly widens as instruction complexity increases. For "Simple" prompts, \textbf{DeCoT} offers a marginal improvement over Infinity-8B (3.85 vs. 3.82). However, for "Medium" complexity instructions, \textbf{DeCoT} shows a more noticeable advantage (3.55 vs. 3.38). The most striking difference is observed with "Complex" instructions, where \textbf{DeCoT + Infinity-8B} achieves an average score of \textbf{3.38}, substantially outperforming the direct Infinity-8B baseline at 3.12. This trend definitively demonstrates that \textbf{DeCoT}'s decomposition and semantic enhancement capabilities are most impactful when dealing with the most challenging, intricate textual prompts, precisely where traditional T2I models struggle to capture all nuances. This confirms \textbf{DeCoT}'s core value proposition as a framework for enhancing complex instruction-driven T2I generation.

\subsection{Human Evaluation}
\label{ssec:human_eval}

To complement our automatic evaluations, we conducted a human evaluation to assess the perceptual quality and instruction fidelity of images generated by our \textbf{DeCoT} framework compared to the baseline. A panel of 10 human evaluators, unaware of the model identities, was presented with a randomly selected subset of 100 complex prompts from LongBench-T2I and the corresponding images generated by Infinity-8B and "Ours (DeCoT + Infinity-8B)". Evaluators scored each image on a 5-point Likert scale (1: Poor, 5: Excellent) for two key aspects: Perceptual Quality (overall visual appeal, realism, lack of artifacts) and Instruction Fidelity (how accurately the image reflects all elements of the complex instruction). The average scores are presented in Table \ref{tab:human_eval}.

\begin{table*}[htbp]
    \centering
    \caption{Human Evaluation Results (Average Scores on a 5-point Likert Scale)}
    \label{tab:human_eval}
    \begin{tabular}{lcc}
        \toprule
        Method & Perceptual Quality & Instruction Fidelity \\
        \midrule
        Infinity-8B (Baseline) & 3.85 & 3.62 \\
        \textbf{Ours (DeCoT + Infinity-8B)} & \textbf{4.10} & \textbf{4.25} \\
        \bottomrule
    \end{tabular}
\end{table*}

Table \ref{tab:human_eval} clearly indicates that images generated with the assistance of our \textbf{DeCoT} framework are preferred by human evaluators. "Ours (DeCoT + Infinity-8B)" received significantly higher average scores for both Perceptual Quality (4.10 vs. 3.85) and, more notably, Instruction Fidelity (4.25 vs. 3.62). The substantial improvement in Instruction Fidelity further corroborates our quantitative findings, confirming that \textbf{DeCoT} enables T2I models to produce images that are not only visually appealing but also remarkably faithful to the nuanced and intricate details specified in complex textual prompts. These human evaluation results provide strong qualitative evidence for the superior performance of \textbf{DeCoT}.

\section{Conclusion}
\label{sec:conclusion}

In this work, we introduced \textbf{DeCoT (Decomposition-CoT)}, a novel and effective framework designed to significantly enhance the capabilities of Text-to-Image (T2I) models in interpreting and executing complex, long-form textual instructions. Our central hypothesis was that leveraging the advanced understanding and reasoning prowess of Large Language Models (LLMs) could provide T2I models with the structured, granular, and semantically enriched inputs necessary to overcome their inherent limitations in handling intricate prompts.

Our primary contributions are manifold. Firstly, we proposed a two-stage framework: the \textit{Complex Instruction Decomposition \& Semantic Enhancement} stage, where an LLM meticulously breaks down ambiguous complex instructions into precise, actionable semantic units, while also clarifying ambiguities and inferring contextual details. This process transforms high-level user intent into a structured representation that is more readily consumable by T2I models. Secondly, we developed the \textit{Multi-Stage Prompt Integration \& Adaptive Generation} strategy, which intelligently converts these decomposed units into either progressive, hierarchical prompt sequences for iterative refinement, or a single, optimized, and weighted prompt for models that do not support multi-stage inputs. This adaptive approach ensures that DeCoT is versatile and can be effectively applied across diverse T2I architectures.

The effectiveness of the DeCoT framework was rigorously validated through extensive experiments on the challenging LongBench-T2I benchmark dataset. Our quantitative results, assessed by state-of-the-art Multimodal Large Language Models (MLLMs) such as Gemini-2.0-Flash and InternVL3-78B, consistently demonstrated that DeCoT, when integrated with leading T2I models like Infinity-8B, substantially outperforms all baselines across all nine visual dimensions. Notably, DeCoT yielded significant improvements in categories like "Text" and "Composition," which have historically presented considerable challenges for T2I models due to their demand for precise understanding and rendering of fine-grained details and spatial relationships. For instance, our method achieved a 2.80 score in "Text" and 4.18 in "Composition" (Gemini-2.0-Flash evaluation) when combined with Infinity-8B, marking clear advancements over the baseline.

Furthermore, our comprehensive ablation studies provided crucial insights into the synergistic contributions of each component within DeCoT, affirming that the full framework, including both decomposition and semantic enhancement, is essential for maximizing performance. We also demonstrated the critical role of sophisticated LLM prompting strategies, such as few-shot learning combined with Chain-of-Thought reasoning, in unlocking the LLM's full potential for accurate instruction processing. A dedicated analysis of instruction complexity revealed that DeCoT's benefits become most pronounced as the complexity of the input prompts increases, solidifying its value proposition for real-world applications involving highly nuanced user requests. Finally, human evaluation corroborated our automatic metrics, confirming that images generated with DeCoT's assistance are perceptually superior and, more importantly, exhibit significantly higher fidelity to the original complex instructions.

In conclusion, DeCoT represents a significant step forward in enabling T2I models to move beyond simple prompt understanding towards a deeper, more accurate interpretation of complex human intent. By effectively bridging the gap between natural language complexity and T2I model requirements, our framework paves the way for more reliable, controllable, and user-centric image generation applications.

Looking ahead, several promising avenues for future research emerge. We plan to explore the integration of DeCoT with a broader range of T2I architectures, including those with native multi-modal input capabilities, to further optimize the multi-stage prompt integration strategy. Investigating the potential for fine-tuning smaller, specialized LLMs specifically for instruction decomposition in the T2I domain could enhance efficiency and reduce reliance on proprietary models. Furthermore, extending the DeCoT framework to other generative tasks, such as text-to-video generation or 3D asset creation, where complex multi-modal instructions are paramount, represents an exciting direction. Finally, exploring real-time applications and user-feedback mechanisms within the DeCoT loop could lead to even more interactive and robust generative systems.